\documentclass[11pt,letterpaper]{article}
\usepackage{ijcnlp2017}
\usepackage{times}
\usepackage{latexsym}

\usepackage{listings}

\ijcnlpfinalcopy

\def\ijcnlppaperid{0}

\title{A parallel corpus of Python functions and documentation strings for automated code documentation and code generation}

\author{Antonio Valerio Miceli Barone \and Rico Sennrich \\
  {\tt amiceli@inf.ed.ac.uk},   {\tt rico.sennrich@ed.ac.uk} \\
  The University of Edinburgh}

\date{}

\begin{document}

\maketitle

\begin{abstract}
Automated documentation of programming source code and automated code generation from natural language are challenging tasks of both practical and scientific interest.
Progress in these areas has been limited by the low availability of parallel corpora of code and natural language descriptions, which tend to be small and constrained to specific domains.

In this work we introduce a large and diverse parallel corpus of a hundred thousands Python functions with their documentation strings ("docstrings") generated by scraping open source repositories on GitHub.
We describe baseline results for the code documentation and code generation tasks obtained by neural machine translation.
We also experiment with data augmentation techniques to further increase the amount of training data.
We release our datasets and processing scripts in order to stimulate research in these areas.
\end{abstract}

\section{Introduction}
\label{SEC:INTRO}

Joint processing of natural languages and programming languages is a research area concerned with tasks such as automated source code documentation, automated code generation from natural language descriptions and code search by natural language queries.
These tasks are of great practical interest, since they could increase the productivity of programmers, and also of scientific interest due to their difficulty and the conjectured connections between natural language, computation and reasoning \cite{Chomsky1956, Miller2003, Graves2014}.


\subsection{Existing corpora}

Major breakthroughs have been recently achieved in machine translation and other hard natural language processing tasks by using neural networks, such as sequence-to-sequence transducers \cite{Bahdanau2014}.
In order to properly generalize, neural networks need to be trained on large and diverse datasets.

These techniques have also been applied with some success to code documentation \cite{Iyer2016} and code generation \cite{Ling2016, Yin2017}, but these works trained and evaluated their models on datasets which are small or limited to restricted domains, in some cases single software projects.

Source code can be collected by scraping open source repositories from code hosting services such as GitHub\footnote{\url{github.com}} \cite{Allamanis2013, Bhoopchand2016}, but the main difficulty is finding natural language annotations that document the code in sufficient detail.

Some existing corpora, such as the the DJANGO dataset and the Project Euler dataset \cite{Oda2015} have been created by human annotators, who can produce high accuracy examples, but this annotation process is expensive and relatively slow, resulting in small (from a few hundreds to less than 20,000 examples) and homogeneous datasets.
Other corpora have been assembled from user-generated descriptions matched to code fragments mined from public websites such as StackOverflow\footnote{\url{stackoverflow.com}} \cite{Allamanis2015, Iyer2016} or IFTTT\footnote{\url{ifttt.com}} \cite{Quirk2015}. These datasets can be large ($>100,000$ examples) but often very noisy.
Another approach is to target a very specific domain, namely trading card games (Magic the Gathering and Hearthstone) \cite{Ling2016}, where code is very repetitive and contains a natural language description (the card text) that can be extracted using simple hand-coded rules. Like the human-annotated corpora, these corpora have high accuracy but are small and very domain-specific.

In practice the existing low-noise corpora seem to have drawbacks which cause them to be unusually easy.
The published evaluation scores on these dataset are are surprisingly high even for baseline systems \cite{Oda2015, Yin2017}, with BLEU scores more than twice those of machine translation between natural languages \cite{
iwslt-report}, a task that we would expect to be no more difficult than code documentation or code generation, especially given the much larger amount of available data.

The DJANGO and and Project Euler corpora use pseudo-code rather than true natural language as a code description, resulting in code fragments and descriptions being similar and easy to align. 
The Magic the Gathering and Hearthstone code fragments are repetitive, with most code of an example being either boilerplate or varying in a limited number of ways that correspond to specific keywords in the description.
We conjecture that, as a consequence of these structural properties, these corpora don't fully represent the complexity of code documentation and code generation as typically done by human programmers, and may be thus of limited use in practical applications.

Therefore we identify the need for a more challenging corpus that better represents code and documentation as they occur in the wild.

\subsection{Our proposal}

In this work we seek to address these limitations by introducing a parallel corpus of over a hundred thousands diverse Python code fragments with descriptions written by their own programmers.

The Python programming language allows each source code object to contain a "docstring" (documentation string), which is retained at runtime as metadata. Programmers use docstrings to describe the functionality and interface of code objects, and sometimes also usage examples. Docstrings can be extracted by automatic tools to generate, for instance, HTML documentation or they can be accessed at runtime when running Python in interactive mode.

We propose the use of docstrings as natural language descriptions for code documentation and code generation tasks.
As the main contribution of this work, we release a parallel corpus of Python function declarations, bodies and descriptions collected from publicly available open source repositories on GitHub.

Current approaches to sequence transduction work best on short and ideally independent fragments, while source code can have complex dependencies between functions and classes.
Therefore we only extract top-level functions since they are usually small and relatively self-contained, thus we conjecture that they constitute meaningful units as individual training examples.
However, in order to support research on project-level code documentation and code generation, we annotate each sample with metadata (repository owner, repository name, file name and line number), enabling users to reconstruct dependency graphs and exploit contextual information. 

We train and evaluate baseline neural machine translation systems for the code documentation and the code generation tasks.
In order to support comparisons using different evaluation metrics, we also release the test and validation outputs of these systems.

We additionally release a corpus of Python functions without docstrings which we automatically annotated with synthetic docstrings created by our code documentation system.
The corpora, extraction scripts and baseline system configurations are available online\footnote{ \url{https://github.com/Avmb/code-docstring-corpus}}.

\section{Dataset}
\label{SEC:DATASET}

\subsection{Extraction and preparation}
\label{SEC:DATASET:EXTRACTION}

We used the GitHub scraper\footnote{\url{https://github.com/uclmr/pycodesuggest}} by \newcite{Bhoopchand2016} with default settings to download source code from repositories on GitHub, retaining Python 2.7 code.

We split each top-level function in a declaration (decorators, name and parameters), a docstring (if present) and the rest of the function body. If the docstring is present, the function is included in the main parallel corpus, otherwise it is included in the "monolingual" code-only corpus for which we later generate synthetic docstrings.

We further process the the data by removing the comments, normalizing the code syntax by parsing and unparsing, removing semantically irrelevant spaces and newlines and escaping the rest and removing empty or non-alphanumeric lines from the docstrings.
Preprocessing removes empty lines and decorative elements from the docstrings but it is functionally reversible on the code\footnote{except in the rare cases where the code accesses its own docstring or source code string}.

An example of an extracted function\footnote{based on scikit-learn \cite{scikit-learn}, with docstring shortened for succinctness} is provided in fig. \ref{FIG:EXAMPLE}.

\subsection{Dataset description}
\label{SEC:DATASET:DESCR}

The extraction process resulted in a main parallel corpus of 150,370 triples of function declarations, docstrings and bodies.

We partition the main parallel corpus in a training/validation/test split, consisting of  109,108 training examples, 2,000 validation examples and 2,000 test examples\footnote{total size is smaller than the full corpus due to duplicate example removal}.


The code-only corpus consists of 161,630 pairs of function declarations and bodies.
The synthetic docstring corpus consists of docstrings generated using from the code-only corpus using our NMT code documentation model, described in the next section.

\begin{table}
\small
\centering
\begin{tabular}{|l|rrr|}
\hline \bf Dataset & \bf Examples & \bf Tokens & \bf LoCs \\ \hline
Parallel decl. & 150,370 & 556,461 & 167,344 \\
Parallel bodies & 150,370 & 12,601,929 & 1,680,176 \\
Parallel docstrings & 150,370 & 5,789,741 & - \\
\hline
Code-only decl. & 161,630 & 538,303 & 183,935 \\
Code-only bodies & 161,630 & 13,009,544 & 1,696,594 \\
\hline
\end{tabular}
\caption{\label{TABLE:CORPUS_STATS_COUNTS} Number of examples, tokens and lines of code in the corpora.}
\end{table}

\begin{table}
\small
\centering
\begin{tabular}{|ll|rrr|}
\hline \bf Corpus & \bf Element & \bf Mean & \bf Std. & \bf Median \\ \hline
Parallel & Declarations & 3.70 & 7.62 & 3 \\
Parallel & Bodies & 83.81 & 254.47 & 40 \\
Parallel & Docstrings & 38.50 & 71.87 & 16 \\
\hline
Code-only & Declarations & 3.33 & 5.04 & 2 \\
Code-only & Bodies & 80.49 & 332.75 & 37 \\
\hline
\end{tabular}
\caption{\label{TABLE:CORPUS_STATS_FREQS} Tokens per example statistics.}
\end{table}

We report corpora summary statistics in tables \ref{TABLE:CORPUS_STATS_COUNTS} and \ref{TABLE:CORPUS_STATS_FREQS}.

\begin{figure*}
\lstset{frame=single, basicstyle=\footnotesize, breaklines=true}
\small
\centering
\begin{tabular}{l}

\begin{lstlisting}
def _intercept_dot(w, X, y):
    """Computes y * np.dot(X, w).
    It takes into consideration if the intercept should be fit or not.
    Parameters
    ----------
    w : ndarray, ndarray, shape (n_features,) or (n_features + 1,)
        Coefficient vector.
    [...]
    """
    c = 0.
    if w.size == X.shape[1] + 1:
        c = w[-1]
        w = w[:-1]
    z = safe_sparse_dot(X, w) + c
    yz = y * z
    return w, c, yz
\end{lstlisting} \\

\begin{lstlisting}
def _intercept_dot(w, X, y):
\end{lstlisting} \\

\begin{lstlisting}
'Computes y * np.dot(X, w). DCNL It takes into consideration if the intercept should be fit or not. DCNL Parameters DCNL w : ndarray, shape (n_features,) or (n_features + 1,) DCNL Coefficient vector. DCNL [...]'
\end{lstlisting} \\

\begin{lstlisting}
 DCSP c = 0.0 DCNL DCSP if (w.size == (X.shape[1] + 1)): DCNL DCSP  DCSP c = w[(-1)] DCNL DCSP  DCSP w = w[:(-1)] DCNL DCSP z = (safe_sparse_dot(X, w) + c) DCNL DCSP yz = (y * z) DCNL DCSP return (w, c, yz)
\end{lstlisting} \\

\begin{lstlisting}
github/scikit-learn/scikit-learn/sklearn/linear_model/logistic.py 39
\end{lstlisting} \\

\end{tabular}
\caption{A Python function with its extracted declaration, docstring, body and repository metadata.}
\label{FIG:EXAMPLE}
\end{figure*}

\section{Baseline results}
\label{SEC:BASELINE}

Since we are releasing a novel dataset, it is useful to assess its difficulty by providing baseline results for other researchers to compare to and hopefully improve upon.

\subsection{Setup}
\label{SEC:BASELINE:SETUP}

In order to obtain these baseline results, we train Neural Machine Translation (NMT) models in both direction using Nematus\footnote{\url{https://github.com/rsennrich/nematus}} \cite{nematus}. 
Our objective here is not to compete with syntax-aware techniques such as \newcite{Yin2017} but to assess a lower bound on the task performance on this dataset without using knowledge of the structure of the programming language.

We prepare our datasets considering the function declarations as part of the input for both the documentation and generation tasks.
In order to reduce data sparsity, we sub-tokenize with the Moses \cite{Koehn2007} tokenization script (which splits some source code identifiers that contain punctuation) followed by Byte-Pair Encoding (BPE) \cite{DBLP:journals/corr/SennrichHB15}. BPE sub-tokenization has been shown to be effective for natural language processing, and for code processing it can be considered a data-driven alternative to the heuristic identifier sub-tokenization of \newcite{Allamanis2015a}.
We train our models with the Adam optimizer \cite{DBLP:journals/corr/KingmaB14} with learning rate $10^{-4}$, batch size $20$.
We use a vocabulary size of $89500$ tokens and we cap training sequence length to $300$ tokens for both the source side and the target side.
We apply "Bayesian" recurrent dropout \cite{Gal2016} with drop probability $0.2$ and word drop probability $0.1$.
We perform early stopping by computing the likelihood every $10000$ on the validation set and terminating when no improvement is made for more than $10$ times.
For the code documentation task, we use word embedding size $500$, state size $500$ and no backpropagation-through-time gradient truncation.
For the code generation task, we use word embedding size $400$, state size $800$ and BPTT gradient truncation at $200$ steps.
These differences are motivated by GPU memory considerations.

After training the code documentation model, we apply it to the corpus-only datasets to generate synthetic docstrings.
We then combine this semi-synthetic corpus to the main parallel corpus to train another code generation model, with the same hyperparameters as above, according to the backtranslation approach of \newcite{2015arXiv151106709S}.

\subsection{Results}
\label{SEC:BASELINE:RESULTS}

\begin{table}
\small
\centering
\begin{tabular}{|l|rr|}
\hline
\bf System & \multicolumn{2}{|c|}{\bf BLEU} \\
 & valid. & test \\
Code-to-docstring & 14.03 & 13.84 \\
Docstring-to-code (base) & 10.32 & 10.24 \\
Docstring-to-code (backtransl.) & 10.85 & 10.90 \\
\hline
\end{tabular}
\caption{\label{TABLE:RESULTS} Code documentation and code generation accuracy (\texttt{multi-bleu.perl}).}
\end{table}

We report BLEU scores for our models in table \ref{TABLE:RESULTS}. Backtranslation provides a moderate improvement of $0.5 - 0.6$ BLEU points over the base model.

Both tasks on this dataset appear to be very challenging, in comparison with the previously published results in the $60-85$ BLEU range by \newcite{Oda2015} and \newcite{Yin2017} on other Python corpora (DJANGO and Hearthstone), which are unusually high compared to machine translation between natural languages, where reaching $40$ BLEU points is challenging.
While BLEU is only a shallow approximation of model accuracy, these large differences are sufficient to demonstrate the challenging nature of our dataset compared to the existing datasets.
We conjecture that this indicative of the strength of our dataset at representing the true complexity of the tasks.

\section{Conclusions}
\label{SEC:CONCLUSIONS}

We argue that the challenging nature of code documentation and code generation is not well represented by the existing corpora because of their drawbacks in terms of noise, size and structural properties.

We introduce a large and diverse parallel corpus of Python functions with their docstrings scraped from public repositories.
We report baseline results on this dataset using Neural Machine Translation, noting that it is much more challenging than previously published corpora as evidenced by translation scores.
We argue that our corpus better captures the complexity of code documentation and code generation as done by human programmers and may enable practical applications.
We believe that our contribution may stimulate research in this area by promoting the development of more advanced models that can fully tackle the complexity of these tasks.
Such models could be, for instance, integrated into IDEs to provide documentation stubs given the code, code stubs given the documentation or context-aware autocomplete suggestions.
Finally, we hope that this research area eventually improves the understanding and possible replication of the human ability to reason about algorithms.

\bibliography{code-docstring-corpus}
\bibliographystyle{ijcnlp2017}

\end{document}